\documentclass[runningheads]{llncs}

\usepackage{graphicx}
\usepackage{color}
\usepackage{etoolbox}
\makeatletter
\patchcmd{\@makecaption}
  {\scshape}
  {}
  {}
  {}
\makeatletter
\patchcmd{\@makecaption}
  {\\}
  {.\ }
  {}
  {}
\makeatother

\usepackage{amsmath,amssymb,amsfonts}
\usepackage{algorithmic}
\usepackage{cite}
\usepackage{float}
\usepackage{subfig}
\usepackage{textcomp}
\usepackage{booktabs}
\usepackage{multirow}
\usepackage{xcolor}
\usepackage{hyperref}
\usepackage{enumitem}
\usepackage{graphicx}
\usepackage[english]{babel}

\setlist[description]{leftmargin=\parindent,labelindent=\parindent}

\usepackage{pgfplots}
\newcommand{\repeatthanks}{\textsuperscript{\thefootnote}}

\begin{document}

\title{Refining Joint Text and Source Code Embeddings for Retrieval Task with Parameter-Efficient Fine-Tuning}

\titlerunning{Refining Joint Text and Code Embeddings for Retrieval with PEFT}

\author{Karim Galliamov\thanks{Equal contribution. Author ordering determined by a dice roll.} \and
Leila Khaertdinova\repeatthanks \and
Karina Denisova\repeatthanks 
}

\authorrunning{Karim Galliamov \and
Leila Khaertdinova \and
Karina Denisova}

\institute{
Innopolis University, Innopolis, Russia \\
\tt\small \{k.galliamov,l.khaertdinova,k.denisova\}@innopolis.university}

\maketitle

\begin{abstract}
The latest developments in Natural Language Processing (NLP) have demonstrated remarkable progress in a code-text retrieval problem. As the Transformer-based models used in this task continue to increase in size, the computational costs and time required for end-to-end fine-tuning become substantial. This poses a significant challenge for adapting and utilizing these models when computational resources are limited. Motivated by these concerns, we propose a fine-tuning framework that leverages Parameter-Efficient Fine-Tuning (PEFT) techniques. Moreover, we adopt contrastive learning objectives to improve the quality of bimodal representations learned by transformer models. Additionally, for PEFT methods we provide extensive benchmarking, the lack of which has been highlighted as a crucial problem in the literature. Based on the thorough experimentation with the CodeT5+ model conducted on two datasets, we demonstrate that the proposed fine-tuning framework has the potential to improve code-text retrieval performance by tuning only 0.4\% parameters at most.

\end{abstract}

\section{Introduction}

The advent of Large Language Models (LLMs) based on Transformer architecture \cite{vaswani2017attention} has revolutionized the field of NLP, offering unprecedented capabilities for understanding and generating human-like text. In the domain of software engineering, these advancements have paved the way for the development of tools that can interpret Natural Language (NL) queries to retrieve the corresponding source code. These tasks hold a significant promise for the development of various Programming Languages (PLs) for both novice and experienced engineers. 

The effective retrieval of source code from NL descriptions, also known as the code search problem, remains challenging, particularly due to the bimodal nature of the task. This problem requires an LLM that can understand and bridge the semantic gap between NL descriptions and PL code. In our work, we emphasize applications of bimodal models \cite{khosla2020supervised} and highlight the notable achievements of pre-trained LLMs applied for code retrieval and generation tasks. End-to-end fine-tuning is a commonly used approach to adapt pre-trained models for a specific task. However, in certain scenarios, especially when applied to smaller datasets, this process can become resource-intensive and overparameterized, leading to minimal or negligible improvements in performance \cite{wang2023codet5}. In addition, overfitting becomes a concern, necessitating the implementation of additional strategies to mitigate this issue. The primary goal of this study is to improve the quality of bimodal representations learned by small transformer models in low-resource settings. In particular, we exploit contemporary Parameter-Efficient Fine-Tuning (PEFT) techniques and contrastive learning to reach performance levels achieved by larger models. This can be reformulated as finding a tradeoff between computational load, caused by an extensive number of trainable parameters, and high resulting performance for code retrieval downstream task. 

Our approach builds upon the capabilities of the CodeT5+ model \cite{wang2023codet5}, utilizing open-source weights, and incorporates a contrastive learning objective to refine the embeddings. Contrastive learning aims to align representations between corresponding text-code instances by maximizing their similarity in the latent space. Furthermore, four contemporary PEFT techniques have been evaluated on two datasets containing nine PLs during this contrastive fine-tuning in low-resource settings. Additionally, we addressed the limitations identified in previous research \cite{lialin2023scaling}, specifically the lack of comprehensive benchmarking for PEFT methods, by providing checkpoints and experiments for our fine-tuned models.

The main contributions of this study are listed as follows:
\begin{itemize}
    \item We incorporate a contrastive learning objective to refine the embeddings on CodeT5+. This allows us to form relevant NL and PL pairs, along with random negatives, improving the baseline model performance.
    \item We introduce an open-source framework \footnote{All final checkpoints can be found in the project repository: \href{https://github.com/leiluk1/CodeSearcher}{https://github.com/leiluk1/CodeSearcher}.} for fine-tuning Transformer encoders, applying PEFT methods for bimodal retrieval tasks. By utilizing PEFT methods, such as LoRA  \cite{yu2023low}, AdaLoRA \cite{zhang2023adaptive}, Prompt-Tuning \cite{lester2021power}, and (IA)3 \cite{liu2022few}, we overcome resource limitations and effectively fine-tune the models for each PL. Besides, we address the limitations identified in \cite{lialin2023scaling} by providing comprehensive benchmarking for PEFT methods. This includes the provision of checkpoints and benchmarks for fine-tuned models.
    \item We evaluate a self-assembled dataset as a Proof of Concept (PoC) and the widely used CodeSearchNet (CSN) benchmark \cite{husain2019codesearchnet} to demonstrate the effectiveness of our approach. This provides a comprehensive assessment of our solution's performance across various PLs, including Python, C++, C\#, SQL, Javascript, Java, Ruby, Go, and PHP.
    \item We integrate our fine-tuned checkpoints into the Retrieval-Augemented Generation (RAG) \cite{lewis2021retrievalaugmented} pipeline as codebase documents and query encoder. This results in a 0.5\% improvement of the ROUGE score for code generation.
\end{itemize}

\section{Related Work}

\subsection{Code-text Retrieval} 

Efficient source code retrieval has been a major area of research, and multiple methods have been explored to bridge the gap between NL queries and PL code \cite{chen2018neural}. A crucial aspect of this field is the use of embeddings to represent code in a manner that enables rapid and accurate retrieval based on semantic similarities \cite{xie2023survey}.

In past years, particularly before the advent of Transformer-like architectures, code retrieval approaches relied on a combination of probabilistic models and classical information retrieval approaches \cite{jiang2016rosf, chatterjee2009sniff, hill2014nl, allamanis2015bimodal}. However, in more recent works, significant breakthroughs have occurred leading to advancements in research in this domain. Notably, the introduction of Transformer-based neural architectures, such as CodeBERT \cite{feng2020codebert}, has revolutionized the field. CodeBERT, an adaptation of the BERT model specifically for programming languages, has set a benchmark for subsequent models in terms of understanding, generating, and retrieving PL codes. To enhance this approach, GraphCodeBERT \cite{guo2020graphcodebert} introduces data flow during pre-training, which effectively captures a semantic-level structure of code. A comparison of CodeBERT with graph-based embeddings for source code representation was presented in \cite{romanov2023comparison}.

Another noteworthy approach is the use of Abstract Syntax Trees (ASTs), which provide a structured representation of the code that captures its syntactic features, facilitating more discerning retrieval and search capabilities \cite{zhang2019novel, liu2020retrieval}. UniSBT, for instance, utilizes syntax-aware embeddings derived from ASTs to enhance the relevance of the retrieved code snippets \cite{gu2021codesearch}.

CodeT5+ extends the T5 model to handle code intricacies, offering improvements in both code understanding, retrieval, and generation tasks \cite{wang2023codet5}. Finally, OpenAI introduced large cpt-code models (from 0.3 to 175 billion parameters) that are pre-trained from scratch using contrastive learning \cite{neelakantan2022text}. While these models have shown groundbreaking performance on code retrieval tasks, it is important to acknowledge that fully pre-training such models with the use of large batch sizes necessitates enormous computational resources and time.

Despite these advancements, the field continues to evolve, with ongoing research seeking to refine these models further and address the challenges posed by the diversity of programming languages and the complexity of code semantics. The work presented in this paper builds upon these foundational efforts, aiming to refine metalanguage embeddings for the retrieval task in low-resource settings through PEFT methods and contrastive learning.

\subsection{Parameter Efficient Fine-Tuning} 
In recent years, the number of parameters in Transformer-based models used in NLP has been growing from millions to trillions. Thus, fine-tuning the parameters of such large models requires substantial computational resources and an enormous amount of time. To overcome these challenges, specific techniques referred to as PEFT have been introduced. In general, these methods allow training a relatively small number of additional parameters in the model, thereby striking a balance between fine-tuning under limited resource scenarios and enabling effective learning of task-specific parameters \cite{lialin2023scaling, xu2023parameter}. For instance, in the context of automated code generation, \cite{weyssow2023exploring} investigates the application of PEFT methods. 

Prompt-tuning is one such approach that introduces additional learnable prompt tokens into the model input. During fine-tuning, only the prompt parameters are updated, whereas the pre-trained parameters remain fixed \cite{lester2021power}. Additionally, Wang et al. \cite{wang2022no} have applied Prompt-tuning to code-related tasks and demonstrated its superiority over fine-tuning models like CodeT5 \cite{wang2021codet5} and CodeBERT \cite{feng2020codebert} in various tasks including code summarization and code translation. Another notable technique, Infused Adapter ((IA)3) \cite{liu2022few} utilizes learned vectors to scale activations, introducing a small number of extra parameters. Additionally, \cite{yu2023low} presents Low-Rank Adaptation (LoRA) that introduces two trainable low-rank matrices for weight update. Building upon the LoRA approach, Adaptive Low-Rank Adaptation (AdaLoRA) \cite{zhang2023adaptive} extends the technique by dynamically adjusting the rank of the matrices to control the allocation budget.

In our study, we explore the usage of all the PEFT techniques described above in combination with contrastive learning to address the source code retrieval task.

\subsection{Contrastive Learning} 

Over the last decade, a plethora of Self-Supervised Learning (SSL) methods have been proposed to learn deep representations without using annotated data \cite{liu2021self}. One of the techniques that has shown state-of-the-art performance in various research fields is contrastive learning. Contrastive learning has been extensively utilized to align representations between different modalities and views in the related literature \cite{radford2021learning, tian2020contrastive, you2020graph, brinzea2022contrastive}. The main idea behind this family of approaches is to maximize the alignment between semantically similar instances by contrasting them against dissimilar ones \cite{chen2020simple}. Specifically, a common way to formulate a contrastive learning objective is to group instances into pairs, positive and negative, and maximize similarities between positive ones. In these settings, a positive pair is formed by different views or modalities corresponding to the same instance.

Initially, contrastive learning has been suggested as an SSL pre-training strategy for deep neural networks. Nevertheless, recent literature presents frameworks that incorporate contrastive losses to fine-tune large models on certain tasks, including information retrieval. For instance, Pour and Farinneya et al. \cite{abdollah2023self} exploited contrastive learning to fine-tune BERT embeddings for retrieving relevant items based on their reviews. In \cite{luo2022clip4clip}, a similar idea, based on CLIP architecture \cite{radford2021learning}, has been proposed for video-text retrieval tasks. Building upon these findings, our study explores fine-tuning the CodeT5+ model \cite{wang2023codet5} using contrastive learning for source code retrieval given textual descriptions.

\section{Methodology}

\subsection{Fine-tuning Approach} 

\begin{figure}[!t] 
    \centering
    \includegraphics[width=12cm]{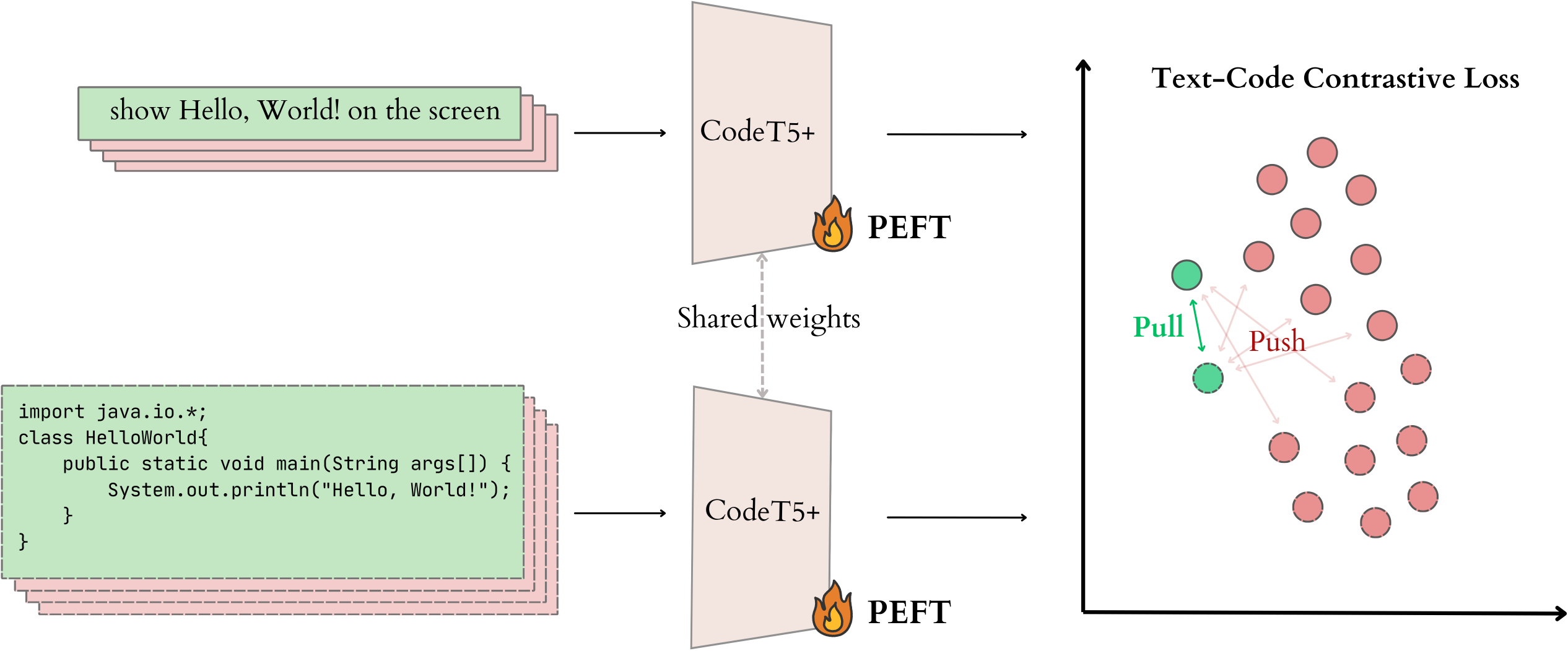}
    
    \caption{The proposed fine-tuning framework. Contrastive loss aims to maximize similarities between corresponding code-text pairs and minimize the similarities of non-matching pairs. For visual clarity, that is schematically demonstrated for one positive pair in a batch, namely "hello world" text and Java code pair. During fine-tuning, CodeT5+ is tuned using PEFT techniques.}
    \label{figure:peft}
\end{figure}

The proposed fine-tuning framework, depicted in Fig. \ref{figure:peft}, utilizes the contrastive learning objective and Parameter-Efficient Fine-tuning techniques. Specifically, we propose to align representations of matching code-text pairs in a joint feature space by utilizing contrastive loss, which is further elaborated on in Section \ref{sec:meth_contrastive}. Furthermore, our contrastive learning approach fine-tunes the CodeT5+ embedding model (110 million parameters) using PEFT methods, that involve training a small proportion of parameters through back-propagation (Section \ref{sec:meth_peft}). An important detail of our fine-tuning is that it is performed for each programming language separately, thus introducing PEFT parameters that adapt the general-purpose CodeT5+ representations for the code retrieval task on a certain programming language. 

\subsection{Contrastive Learning to Align Code-text Pairs} 
\label{sec:meth_contrastive}

In the pursuit of refining source code retrieval from NL descriptions, our methodology emphasizes the alignment of code-text pairs as a critical aspect. This alignment process is instrumental in bridging the semantic gap between NL queries and PL codes, which is essential for effective retrieval systems. In order to achieve this, we have employed PEFT methods on the CodeT5+ model, leveraging its bimodal capabilities to understand and generate code. The training of embeddings differs from generating a PL conditioned on NL input. More specifically, the objective would not be next-token likelihood optimization, but rather bimodal contrastive loss. Contrastive learning is an approach that aims to bring closer the embeddings of relevant NL and PL pairs while pushing apart those of irrelevant pairs. This is achieved by using a text-code contrastive loss function during training. Consequently, the model acquires the ability to generate more discriminative embeddings, thereby facilitating improved alignment between NL queries and their corresponding code.

In our study, contrastive loss will pull together embeddings for positive (relevant) samples and pull apart irrelevant pairs \cite{wang2023codet5}. Formally, given a mini-batch of N code-text pairs $\{\boldsymbol{h}_i^c, \boldsymbol{h}_i^t\}_{i = 1}^N$ as normalized lower-dimensional representations, relevant pairs of code vector $\boldsymbol{h}_i^c$ and text vector $\boldsymbol{h}_i^t$ are considered positive as they correspond to the same $i$-th instance in a batch. The loss $l_i^{t \rightarrow c}$ treating textual representation $\boldsymbol{h}_i^t$ from $i$-th example as an anchor can be computed as follows:

\begin{equation}
   l_i^{c \rightarrow t} = - log \frac{\delta({\boldsymbol{h}}_i^{c},{\boldsymbol{h}}_i^{t})}{\sum_{k=1}^N \delta({\boldsymbol{h}}_i^{c},{\boldsymbol{h}}_k^{t})},
   \label{eq:nt-xent}
\end{equation}

where $\delta({\boldsymbol{h}}_i^{c}{\boldsymbol{h}}_i^{t}) = \exp\left(\frac{{\boldsymbol{h}}_i^{c T} {\boldsymbol{h}}_i^{t}}{\tau}\right)$. Therefore, the total loss aggregated for the whole batch of views $c$ and $t$ can be averaged as:

\begin{equation}
    L^{c,t} = \frac{1}{2N}\sum_{i=1}^N (l_i^{c \rightarrow t} + l_i^{t \rightarrow c})
\end{equation}

\subsection{Trainable Parameters}
\label{sec:meth_peft}

We utilize PEFT methods that involve training a small proportion of newly added parameters through back-propagation. In our setup, during training, all model weights are frozen except for the ones added by PEFT methods. These methods are defined as follows:
\begin{itemize}
    \item LoRA \cite{yu2023low}: introducing low-rank addends to $Q, V$ tensors of Attention.
    \item AdaLoRA \cite{zhang2023adaptive}: LoRA-based approach with addends ranks changing during training.
    \item (IA)3 \cite{liu2022few}: learnable scaling vectors for $Q$, $K$, and linear tensors.
    \item Prompt-Tuning \cite{lester2021power}: learnable vectors prepended to hidden input representation.
\end{itemize}

\section{Experimental setup}

In the course of our research, we have meticulously designed an experimental setup to evaluate the efficacy of PEFT methods applied to the CodeT5+ model for the task of source code retrieval for small models. Our experiments were conducted with a focus on optimizing the alignment of NL and PL embeddings, which is crucial for the successful retrieval of source code corresponding to NL descriptions.

\subsection{Data Selection}
\label{sec:exp_data}

For our fine-tuning framework, we utilized two distinct datasets: the well-established CodeSearchNet benchmark and a custom dataset assembled specifically for this study.

\subsubsection{CodeSearchNet}

The CodeSearchNet (CSN) \cite{husain2019codesearchnet} benchmark is a large-scale dataset that has been widely used in the field for evaluating code retrieval models. This dataset encompasses a diverse range of programming languages, including Java, JavaScript, Go, PHP, and Ruby. This dataset provides a comprehensive set of NL documentation and PL code pairs. Table~\ref{table:CSN_datasets_sizes} demonstrates dataset split sises for each PL. The CSN dataset has been used as a benchmark for demonstrating the robustness and effectiveness of our approach across a wide range of programming languages, allowing us to assess the performance of our fine-tuned models comprehensively.

\begin{table}
    \centering
    \begin{tabular}{ p{1.7cm}p{1.7cm}p{1.7cm}p{1.7cm}p{1.7cm}p{1.7cm} } 
        \hline
        & Ruby   & JS   & Go   & Java & PHP  \\ 
        \hline
        train & 48K & 123K & 317K & 454K & 523K \\
        valid & 2.2K & 8.2K & 14K & 15K & 26K \\
        test & 2.2K & 6.4K & 14K & 26K & 28K \\
        \hline
    \end{tabular}
    \caption{Size of the CodeSearchNet dataset per programming language and split.}
    \label{table:CSN_datasets_sizes}
\end{table}

\subsubsection{Custom Dataset}
To prove our concept and demonstrate the potential of PEFT methods in enhancing the CodeT5+ model's ability to map NL descriptions to source code we collected a relatively small dataset compared to the CodeSearchNet. The custom dataset includes a collection of NL text and PL code pairs, with a particular emphasis on smaller and less represented programming languages, which are neglected in larger benchmarks. 

In our experimental setup, we investigated various PL datasets, including Python \cite{zhu2022xlcost, yao2018staqc, bahrami2021pytorrent}, C\# \cite{zhu2022xlcost}, C++ \cite{zhu2022xlcost}, SQL \cite{rao2021search4code}, Solidity \cite{shi2021semantic}, and Assembly \cite{kairajarvi2020towards}. However, we determined that the Solidity and Assembly datasets were not suitable, as it was impossible to match PL and NL pairs: Assembly dataset did not contain documentation, and Solidity dataset was not publicly available anymore. We therefore excluded these datasets. During the data exploration, a thorough analysis of the selected datasets was conducted to gain insight into their characteristics and identify any potential biases or data quality issues. Several datasets, particularly Search4Code \cite{rao2021search4code}, were removed, as they do not provide any code snippets and NL query required for the task of code search. Moreover, pairs with non-English natural language queries were removed. Additionally, we deleted NL and PL pairs, where each sample in the pair was too short, specifically less than three tokens, or too long, the boundary was defined as a hyperparameter. The absence of appropriate code snippets and related code descriptions restricts our ability to include them in the final dataset. 

\begin{figure}[!ht] 
    \centering
    \includegraphics[width=12cm, height=7cm]{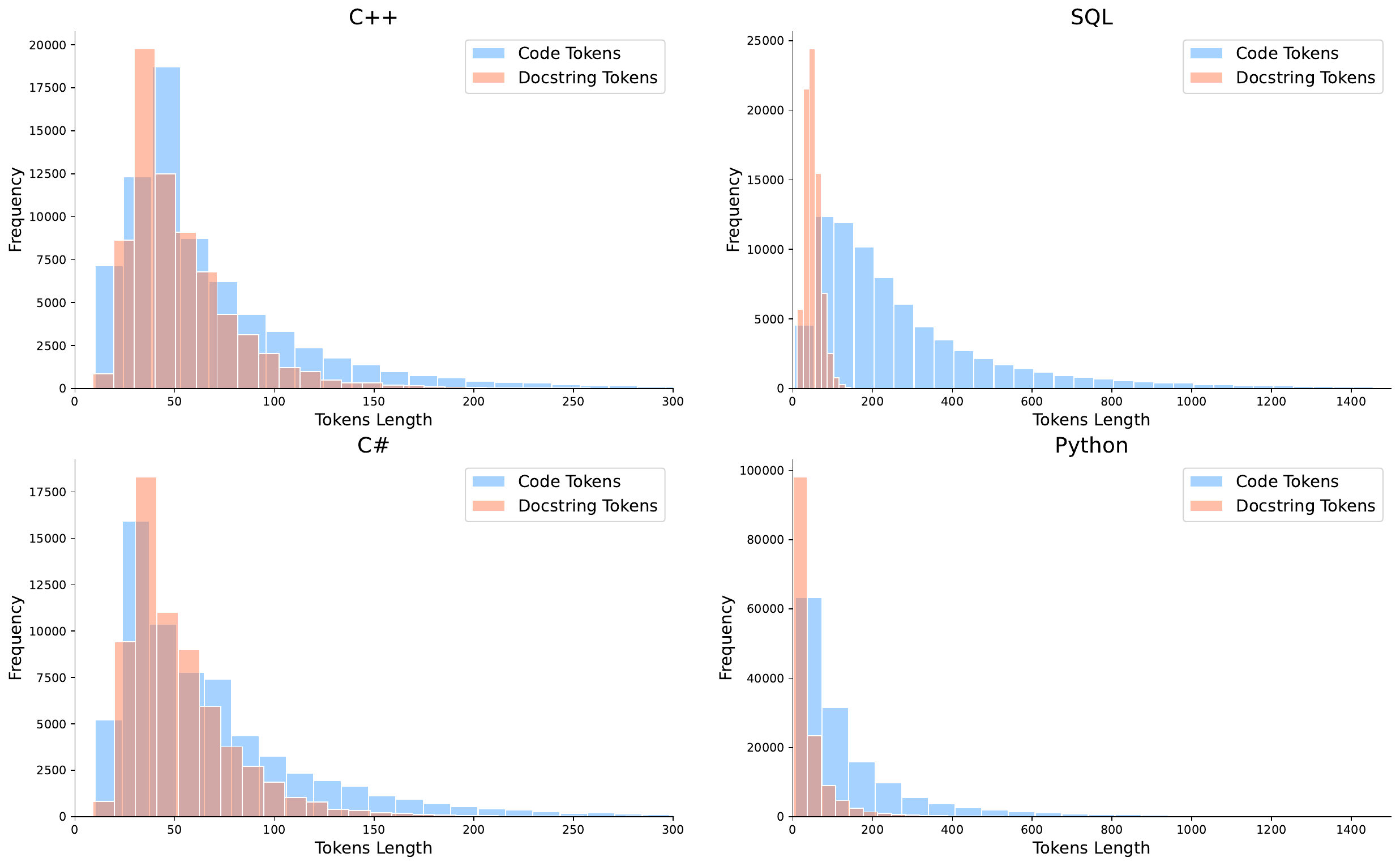}
    
    \caption{The distribution of token length for NL code docstring and PL code snippets, respectively, for PLs included in our dataset.}
    \label{figure:eda}
\end{figure}

Following the exploration of these datasets, we merged them according to their respective programming languages. Then, tokenization and common preprocessing steps were performed for both the NL description and PL code. For further model fine-tuning, we establish the token lengths of NL text and PL code. Figure ~\ref{figure:eda} demonstrates detailed information on the distribution of both the PL and NL token lengths in the dataset.

As a result, we obtained the dataset with sample split sizes demonstrated in Table ~\ref{table:datasets_sizes}. By incorporating these datasets into our experiments, we aimed to showcase the versatility of our fine-tuning framework and its applicability to a wide array of programming languages.

\begin{table}
    \centering
    \begin{tabular}{ p{1.3cm}p{1.3cm}p{1.3cm}p{1.3cm}p{1.3cm} }
        \hline
        & Python  & C\# & C++ & SQL \\ 
        \hline
        train & 10M & 58K & 63K & 62.9K \\ 
        valid & 120K & 3K & 3K & 6.9K  \\
        test & 123K & 5.4K & 5.6K & 7.7K \\
        \hline
    \end{tabular}
    \caption{Size of the custom dataset per programming language and split.}
    \label{table:datasets_sizes}
\end{table}

\subsection{Fine-tuning Details}
For the base pre-trained model we have taken a CodeT5+ encoder\footnote{https://huggingface.co/Salesforce/codet5p-110m-embedding} with a lower dimension projection and normalization head on top. The above-mentioned encoder comes from the CodeT5 model, pre-trained on large-scale datasets on tasks of code generation, span denoising, contrastive objectives, and others. For more details on how the base model was pre-trained, one can refer to \cite{wang2021codet5}.

For fine-tuning, we chose to train on pairs with at most 256 NL and 256 PL maximum-length tokens.
For the embedding loss, the initial temperature $\tau$ was set to 0.08. The learning rates were set to $0.001$, batch size was set to $128$ and the gradient accumulation steps were set to $4$ in all the experiments. Furthermore, a Cosine Annealing Scheduler was used. 

\begin{table}
    \centering
    \begin{tabular}{ p{1.7cm}p{1.7cm}p{1.7cm}p{1.7cm}p{1.7cm}}
        \hline
        & AdaLoRA  & LoRA & \( (IA)3 \) & Prompt \\ 
        \hline
        tunable \% & 0.402 & 0.268 & 0.025 & 0.007 \\ 
        tunable \# & 442,656 & 294,912 & 27,648 & 7680  \\
        \hline
    \end{tabular}
    \caption{Tunable parameters for PEFT methods. Prompt-tuning was done with 10 tokens by default.}
    \label{table:trainparam}
\end{table}

When tuning the embeddings, we used Prompt-tuning instead of Prompt Encoder from the PEFT library. This decision was made because the latter failed to overfit a single batch, and Prefix-Tuning was not supported for the embedding extraction task. 

All final checkpoints may be found in our project repository\footnote{https://github.com/leiluk1/CodeSearcher/tree/main/checkpoints}.

\subsection{Evaluation Methods}

We use Mean Reciprocal Rank (MRR) as the evaluation metric. For the evaluation part of the validation and test sets, we have utilized two approaches to calculate MRR. The first approach, adopted from \cite{feng2020codebert}, involves computing similarities for all possible NL, and PL pairs in the test dataset. The MRR is then calculated only on the ranks that are not greater than $1000$. The second approach, proposed in \cite{husain2019codesearchnet}, splits the test dataset into chucks of $1000$ pairs each (if the last chunk is smaller, then it is discarded). Regular mean reciprocal ranks are computed for each of these chucks, and these MRRs are then averaged. In each table with the evaluation part presented in our work, the chosen method for calculating the MRR is given in the corresponding description.

\subsection{Further Usage with RAG} 
\label{seq:rag_desc}
\begin{figure}[!t] 
    \centering
    \includegraphics[width=12cm, height=7cm]{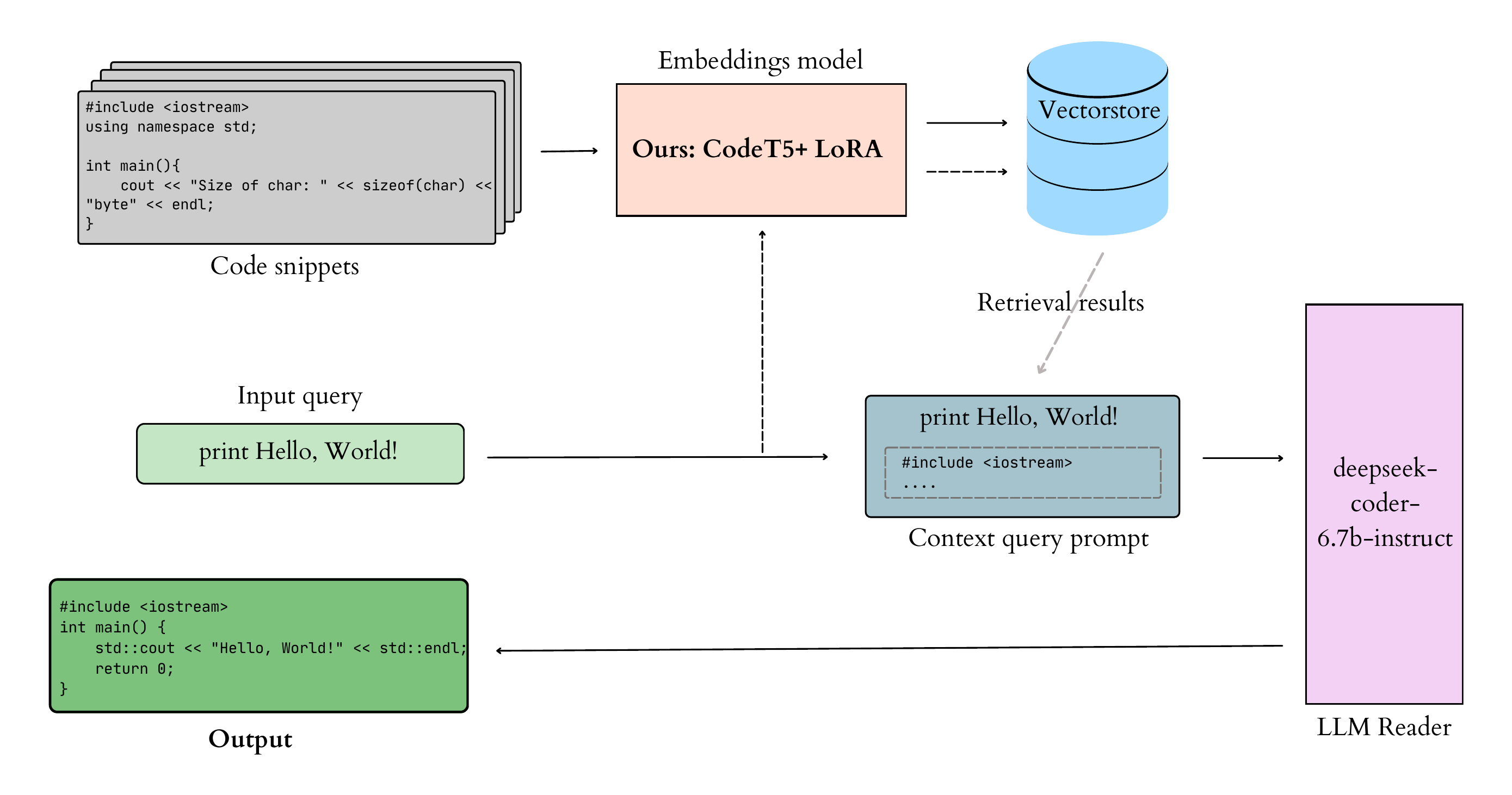}

    \caption{Integration of the best checkpoints of our fine-tuned models into the RAG pipeline for different PLs used in the study. The figure provides an example of the code generation for C++.}
    \label{figure:rag}
\end{figure}

One of the points of application of our models could be Retrieval-Augmented Generation (RAG) \cite{lewis2021retrievalaugmented}. We incorporate our model as an embedding model for both text and code chunks. The retrieval database is built on code samples, and corresponding docstrings are used as queries. For reader LLM, we have chosen "deepseek-coder-6.7b-instruct" from \cite{guo2024deepseekcoder}. To provide a visual representation of our approach in the context of RAG, please refer to Figure \ref{figure:rag}. This resulted in a minor but stable improvement of the ROUGE metric compared to the non-tuned baseline model. See more details about evaluation results in section \ref{sec:res_rag}. 

\section{Results}

\subsection{Custom Dataset}
First, the proposed fine-tuning strategy was applied to the custom dataset described in Section \ref{sec:exp_data}. In particular, we evaluated the suggested contrastive learning objective along with PEFT techniques on each programming language presented in the dataset. Table \ref{table:mrr} summarizes the MRR scores obtained on test sets of the dataset. In this table, we also show the baseline results that correspond to the pre-trained CodeT5+ without any further fine-tuning. According to the obtained results, all the utilized PEFT techniques lead to an increase in performance. Among these methods, AdaLoRA demonstrates the highest scores in all four programming languages. Specifically, it boosts the performance by about 17\% on C++ and C\#, more than 6\% on SQL, and approximately 9\% on Python compared to the baseline scores.

\begin{table}[!ht] 
    \centering
    \begin{tabular}{ p{2.0cm}p{2.0cm}p{2.0cm}p{2.0cm}p{2.0cm}} 
        \hline
        & Python & C\# & C++  & SQL \\ 
        \hline
        Baseline & 75.89 & 22.15 & 21.69 & 06.15 \\ 
        LoRA & 83.22 & 38.75 & 39.23 & 12.57 \\
        AdaLoRA & \textbf{83.95} & \textbf{39.19} & \textbf{39.31} & \textbf{12.80} \\
        \( (IA)3 \) & 78.48 & 32.99 & 32.61 & 08.97 \\
        Prompt & 81.33 & 28.85 & 23.84 & 7.91 \\
        \hline
    \end{tabular}
    \caption{Embeddings model evaluation results: MRR on test datasets (for Python, the number of test pairs was limited to 32k) calculated by the same function as in CodeBert \cite{feng2020codebert}.}
    \label{table:mrr}
\end{table}

\begin{figure}[!ht] 
    \centering
    \subfloat{{\includegraphics[width=12cm, height=7cm]{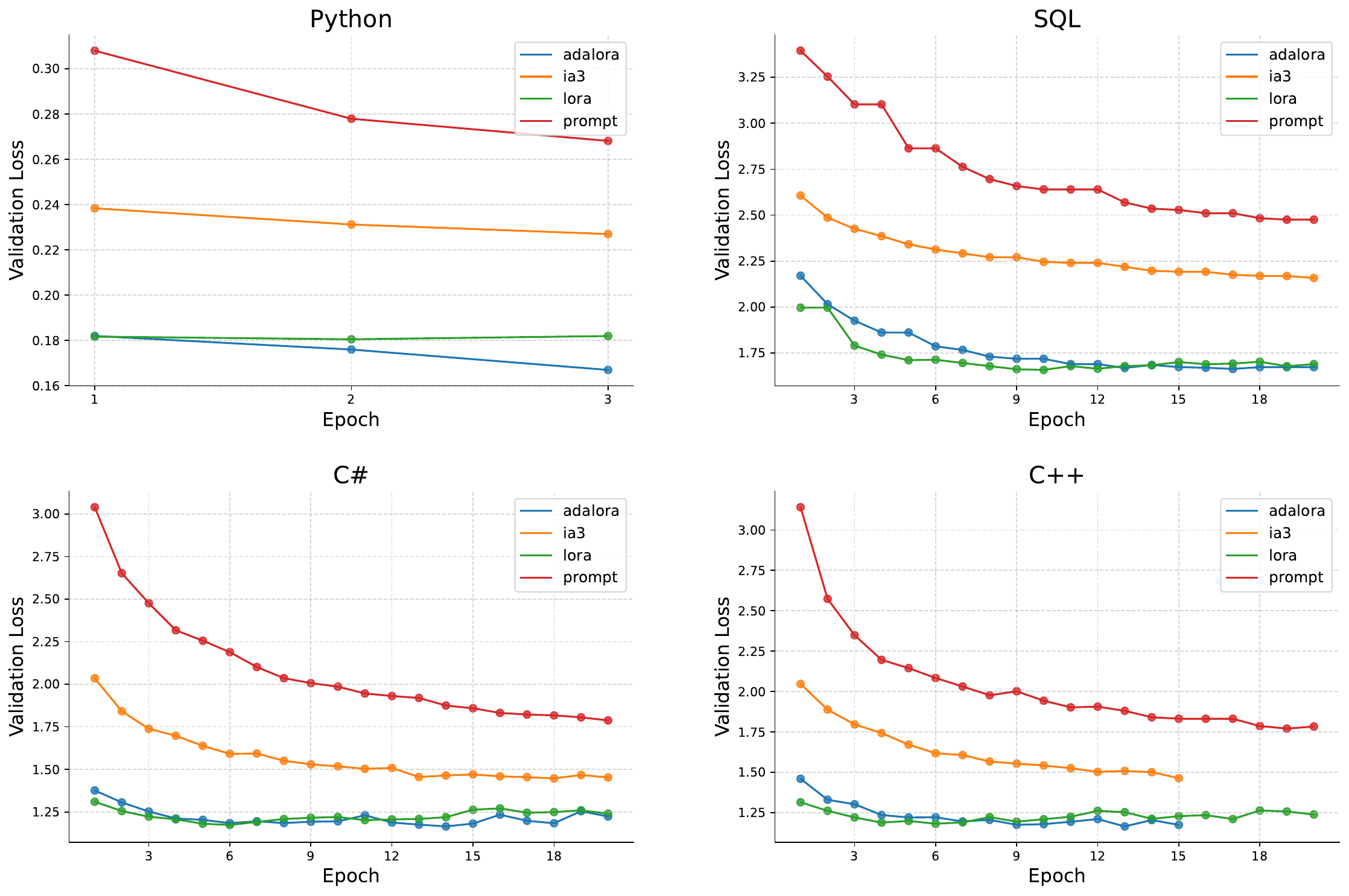}}}
    \caption{Validation losses plots for the embeddings model on PoC datasets.}
    \label{figure:loss}
\end{figure}

Furthermore, Fig. ~\ref{figure:loss} illustrates the validation losses during fine-tuning. Based on the obtained curves a few observations can be made. First, reparametrization-based methods such as LoRA and AdaLoRA seem to have the fastest convergence. (IA)3 converges a little bit slower, while Prompt Tuning is the longest to converge. What is more, for SQL, the loss converges to higher values. This can be explained by the fact that SQL was not used during the pre-training of CodeT5+ \cite{wang2023codet5}, meaning that the model performs in zero-shot settings. Nevertheless, our fine-tuning improves the performance of this language as well, as highlighted above.

Finally, it is worth noticing that due to limited computational resources and the large size of Python data, we conducted fine-tuning for only 3 epochs. However, as mentioned earlier, we still observe a substantial enhancement in performance after our fine-tuning. We leave the exploration of more thorough fine-tuning using Python data for future work.

\subsection{Benchmarking on CSN}

We evaluated our approach on the CSN benchmark described in Section \ref{sec:exp_data}. The results, provided in Table ~\ref{table:mrr_csn}, demonstrate the MRR scores achieved on the test sets of respective PLs within the CSN dataset. Similarly, Table ~\ref{table:csn_comparison} provides a comparison of MRR scores obtained in our approach with those of SOTA models, while also considering the number of trainable parameters.

\begin{table}[!ht] 
    \centering
    \begin{tabular}{ p{1.7cm}p{1.7cm}p{1.7cm}p{1.7cm}p{1.7cm}p{1.7cm} } 
        \hline
        & Ruby   & JS   & Go   & Java & PHP  \\ 
        \hline
        Baseline & 76.11 & 74.42 & 77.69 & 75.66 & 77.39  \\ 
        LoRA   & 77.74 & \textbf{76.99} & \textbf{79.10} & \textbf{77.82} & \textbf{80.43}  \\
        AdaLoRA & \textbf{77.88} & 76.76  & 78.99 & 77.53 & 80.24 \\
        \( (IA)3 \)   & 76.98 & 76.14 & 78.95 & 77.37 & 79.70 \\
        Prompt  & 74.53 & 74.93  & 78.49 & 76.73 & 79.00 \\
        \hline
        
    \end{tabular}
    \caption{Embeddings model evaluation results on CSN benchmark. For MRR calculation, an approach from \cite{husain2019codesearchnet} was used.}
    \label{table:mrr_csn}
\end{table}

Based on results from Table ~\ref{table:mrr_csn}, we observed the greatest improvement over baseline performance when using LoRA and AdaLoRA methods. However, compared to the results on the custom dataset, LoRA slightly outperforms AdaLoRA almost for all programming languages, except Ruby. Selected PEFT methods outperformed the baseline across all programming languages presented in CSN. LoRA achieved the highest MRR JavaScript (+2.5\%), Go (+1.3\%), Java (+2.1\%), and PHP (+3\%) followed closely by AdaLoRA, which performed the best results on Ruby (+1.7\%).  IA3 also showed improvements over the baseline but was less effective than LoRA and AdaLoRA. 

\begin{table}[!ht]
\centering
\begin{tabular}{l|ccccc|c}
\hline
 & \multicolumn{5}{c|}{CodeSearchNet} &  \\ \cline{2-6}
\multirow{-2}{*}{Model} & Ruby & JS & Go & Java & PHP & \multirow{-2}{*}{Trainable parameters} \\ \hline
GraphCodeBERT \cite{guo2020graphcodebert} &   70.3 &   64.4 &   \underline{89.7} &   69.1 &   64.9 & 125M \\
CodeBERT \cite{feng2020codebert} &   67.9 &   62.0 &   88.2 &   67.6 &   62.8 & 125M \\ \hline
Ours: CodeT5+ baseline &   76.1 &   74.4 &   77.7 &   75.7 &   77.4 & 220M \\
Ours: CodeT5+ AdaLoRA & \underline{77.9} &   76.8 &   79.0 &   77.5 &   80.2 & 443k \\
Ours: CodeT5+ LoRA &   77.7 &  \underline{77.0} & 79.1 & \underline{77.8} &  \underline{80.4} &  295k \\ \hline
cpt-code S \cite{neelakantan2022text} & \textbf{86.3} &   86.0 & \textbf{97.7} &   94.0 &   96.7 & 300M \\
cpt-code M  \cite{neelakantan2022text} &   85.5 & \textbf{86.5} &   97.5 & \textbf{94.4} & \textbf{97.2} & 1.2B \\
\hline
\end{tabular}
\caption{Comparison of our approach against state-of-the-art models. We underline the second-best results after cpt-code models. We take the text-to-code retrieval results for other models, except ours and CodeT5+ baseline, from \cite{wang2023codet5} and \cite{neelakantan2022text}. For the Go benchmark, we obtained substantially lower results compared to the original CodeT5+ paper \cite{wang2023codet5} when using the open-source checkpoint for CodeT5+.}
\label{table:csn_comparison}
\end{table}

The findings presented in  ~\ref{table:csn_comparison} highlight the performance of our approach in comparison to SOTA models. Our approach stands out as achieving the second-best results, following the performance of cpt-code models. It is important to note that the cpt-code S and M models, which outperform our approach, are significantly larger with 0.3 and 1.2 billion parameters, respectively. Moreover, all these parameters were tuned during pre-training using contrastive learning in an end-to-end fashion. In addition, we achieve an increase of 6.7\% in average MRR compared to BERT-based embedders. A notable aspect of our approach is the fine-tuning of CodeT5+ using LoRA, which not only contributes to its high performance but also makes it the most advantageous option in terms of computational expenses. This technique finds an optimal balance between the number of trainable parameters and MRR scores, further bridging a gap between CodeT5+ and cpt-code. 

\subsection{RAG Case Study}
\label{sec:res_rag}
Exact settings for RAG setup may be found in the Experimental Setup section ~\ref{seq:rag_desc}.
Computing the ROUGE \cite{ganesan2018rouge} over 1000 docstring queries from CSN test split has given an 0.5\% increase in ROUGE-L, 0.6\% increase in ROUGE-2 and 0.45\% increase in ROUGE-1. It is definite that further experimentation with various reader models and prompts is needed, which lies outside of the scope of our work.

\section{Conclusion}

In this paper, we adopted a contrastive learning objective to enhance source code embeddings for retrieval tasks along with fine-tuning CodeT5+ with PEFT methods in low-resource settings. To address the existing limitation of comprehensive benchmarks for PEFT techniques, we developed an open-source framework for fine-tuning CodeT5+ using PEFT techniques. We evaluated our fine-tuned model on diverse programming language datasets, including our custom dataset and the CodeSearchNet.

Our findings provide a foundation for future studies and highlight the potential of PEFT techniques. However, it is crucial to acknowledge the limitations of our work. In particular, this is related to limited resource power, the small batch size used in fine-tuning using contrastive learning, and the small number of epochs used for large datasets.

In future work, we propose exploring additional PEFT methods, expanding the evaluation on different code datasets, and investigating techniques to handle larger codebases. Another direction of research could be aligning embedder models for specific reader models in the RAG pipeline.

\section*{Acknowledgements}
We would like to express our thanks to Innopolis University for providing part of resources and facilities that were essential for conducting the experiments in this work. We extend our sincere gratitude to Professor V. Ivanov from Innopolis University for his invaluable guidance and support throughout our research. 

\medskip
\bibliographystyle{unsrt}
\bibliography{references}

\begin{thebibliography}{10}

\bibitem{vaswani2017attention}
Ashish Vaswani, Noam Shazeer, Niki Parmar, Jakob Uszkoreit, Llion Jones, Aidan~N Gomez, {\L}ukasz Kaiser, and Illia Polosukhin.
\newblock Attention is all you need.
\newblock {\em Advances in neural information processing systems}, 30, 2017.

\bibitem{khosla2020supervised}
Prannay Khosla, Piotr Teterwak, Chen Wang, Aaron Sarna, Yonglong Tian, Phillip Isola, Aaron Maschinot, Ce~Liu, and Dilip Krishnan.
\newblock Supervised contrastive learning.
\newblock {\em Advances in neural information processing systems}, 33:18661--18673, 2020.

\bibitem{wang2023codet5}
Yue Wang, Hung Le, Akhilesh Gotmare, Nghi Bui, Junnan Li, and Steven Hoi.
\newblock Codet5+: Open code large language models for code understanding and generation.
\newblock In {\em Proceedings of the 2023 Conference on Empirical Methods in Natural Language Processing}, pages 1069--1088, 2023.

\bibitem{lialin2023scaling}
Vladislav Lialin, Vijeta Deshpande, and Anna Rumshisky.
\newblock Scaling down to scale up: A guide to parameter-efficient fine-tuning.
\newblock {\em arXiv preprint arXiv:2303.15647}, 2023.

\bibitem{yu2023low}
Yu~Yu, Chao-Han~Huck Yang, Jari Kolehmainen, Prashanth~G Shivakumar, Yile Gu, Sungho Ryu~Roger Ren, Qi~Luo, Aditya Gourav, I-Fan Chen, Yi-Chieh Liu, et~al.
\newblock Low-rank adaptation of large language model rescoring for parameter-efficient speech recognition.
\newblock In {\em 2023 IEEE Automatic Speech Recognition and Understanding Workshop (ASRU)}, pages 1--8. IEEE, 2023.

\bibitem{zhang2023adaptive}
Qingru Zhang, Minshuo Chen, Alexander Bukharin, Pengcheng He, Yu~Cheng, Weizhu Chen, and Tuo Zhao.
\newblock Adaptive budget allocation for parameter-efficient fine-tuning.
\newblock {\em arXiv preprint arXiv:2303.10512}, 2023.

\bibitem{lester2021power}
Brian Lester, Rami Al-Rfou, and Noah Constant.
\newblock The power of scale for parameter-efficient prompt tuning.
\newblock {\em arXiv preprint arXiv:2104.08691}, 2021.

\bibitem{liu2022few}
Haokun Liu, Derek Tam, Mohammed Muqeeth, Jay Mohta, Tenghao Huang, Mohit Bansal, and Colin~A Raffel.
\newblock Few-shot parameter-efficient fine-tuning is better and cheaper than in-context learning.
\newblock {\em Advances in Neural Information Processing Systems}, 35:1950--1965, 2022.

\bibitem{husain2019codesearchnet}
Hamel Husain, Ho-Hsiang Wu, Tiferet Gazit, Miltiadis Allamanis, and Marc Brockschmidt.
\newblock Codesearchnet challenge: Evaluating the state of semantic code search.
\newblock {\em arXiv preprint arXiv:1909.09436}, 2019.

\bibitem{lewis2021retrievalaugmented}
Patrick Lewis, Ethan Perez, Aleksandra Piktus, Fabio Petroni, Vladimir Karpukhin, Naman Goyal, Heinrich Küttler, Mike Lewis, Wen tau Yih, Tim Rocktäschel, Sebastian Riedel, and Douwe Kiela.
\newblock Retrieval-augmented generation for knowledge-intensive nlp tasks, 2021.

\bibitem{chen2018neural}
Qingying Chen and Minghui Zhou.
\newblock A neural framework for retrieval and summarization of source code.
\newblock In {\em Proceedings of the 33rd ACM/IEEE International Conference on Automated Software Engineering}, pages 826--831, 2018.

\bibitem{xie2023survey}
Yutao Xie, Jiayi Lin, Hande Dong, Lei Zhang, and Zhonghai Wu.
\newblock Survey of code search based on deep learning.
\newblock {\em ACM Transactions on Software Engineering and Methodology}, 33(2):1--42, 2023.

\bibitem{jiang2016rosf}
He~Jiang, Liming Nie, Zeyi Sun, Zhilei Ren, Weiqiang Kong, Tao Zhang, and Xiapu Luo.
\newblock Rosf: Leveraging information retrieval and supervised learning for recommending code snippets.
\newblock {\em IEEE Transactions on Services Computing}, 12(1):34--46, 2016.

\bibitem{chatterjee2009sniff}
Shaunak Chatterjee, Sudeep Juvekar, and Koushik Sen.
\newblock Sniff: A search engine for java using free-form queries.
\newblock In {\em Fundamental Approaches to Software Engineering: 12th International Conference, FASE 2009, Held as Part of the Joint European Conferences on Theory and Practice of Software, ETAPS 2009, York, UK, March 22-29, 2009. Proceedings 12}, pages 385--400. Springer, 2009.

\bibitem{hill2014nl}
Emily Hill, Manuel Roldan-Vega, Jerry~Alan Fails, and Greg Mallet.
\newblock Nl-based query refinement and contextualized code search results: A user study.
\newblock In {\em 2014 Software Evolution Week-IEEE Conference on Software Maintenance, Reengineering, and Reverse Engineering (CSMR-WCRE)}, pages 34--43. IEEE, 2014.

\bibitem{allamanis2015bimodal}
Miltos Allamanis, Daniel Tarlow, Andrew Gordon, and Yi~Wei.
\newblock Bimodal modelling of source code and natural language.
\newblock In {\em International conference on machine learning}, pages 2123--2132. PMLR, 2015.

\bibitem{feng2020codebert}
Zhangyin Feng, Daya Guo, Duyu Tang, Nan Duan, Xiaocheng Feng, Ming Gong, Linjun Shou, Bing Qin, Ting Liu, Daxin Jiang, et~al.
\newblock Codebert: A pre-trained model for programming and natural languages.
\newblock {\em arXiv preprint arXiv:2002.08155}, 2020.

\bibitem{guo2020graphcodebert}
Daya Guo, Shuo Ren, Shuai Lu, Zhangyin Feng, Duyu Tang, Shujie Liu, Long Zhou, Nan Duan, Alexey Svyatkovskiy, Shengyu Fu, et~al.
\newblock Graphcodebert: Pre-training code representations with data flow.
\newblock {\em arXiv preprint arXiv:2009.08366}, 2020.

\bibitem{romanov2023comparison}
Vitaly~Anatolyevich ROMANOV and Vladimir~Vladimirovich IVANOV.
\newblock Comparison of graph embeddings for source code with text models based on cnn and codebert architectures.
\newblock {\em Proceedings of the Institute for System Programming of the RAS (Proceedings of ISP RAS)}, 35(1):237--264, 2023.

\bibitem{zhang2019novel}
Jian Zhang, Xu~Wang, Hongyu Zhang, Hailong Sun, Kaixuan Wang, and Xudong Liu.
\newblock A novel neural source code representation based on abstract syntax tree.
\newblock In {\em 2019 IEEE/ACM 41st International Conference on Software Engineering (ICSE)}, pages 783--794. IEEE, 2019.

\bibitem{liu2020retrieval}
Shangqing Liu, Yu~Chen, Xiaofei Xie, Jingkai Siow, and Yang Liu.
\newblock Retrieval-augmented generation for code summarization via hybrid gnn.
\newblock {\em arXiv preprint arXiv:2006.05405}, 2020.

\bibitem{gu2021codesearch}
Jian Gu, Zimin Chen, and Martin Monperrus.
\newblock Multimodal representation for neural code search.
\newblock In {\em 2021 IEEE International Conference on Software Maintenance and Evolution (ICSME)}, pages 483--494. IEEE, 2021.

\bibitem{neelakantan2022text}
Arvind Neelakantan, Tao Xu, Raul Puri, Alec Radford, Jesse~Michael Han, Jerry Tworek, Qiming Yuan, Nikolas Tezak, Jong~Wook Kim, Chris Hallacy, et~al.
\newblock Text and code embeddings by contrastive pre-training.
\newblock {\em arXiv preprint arXiv:2201.10005}, 2022.

\bibitem{xu2023parameter}
Lingling Xu, Haoran Xie, Si-Zhao~Joe Qin, Xiaohui Tao, and Fu~Lee Wang.
\newblock Parameter-efficient fine-tuning methods for pretrained language models: A critical review and assessment.
\newblock {\em arXiv preprint arXiv:2312.12148}, 2023.

\bibitem{weyssow2023exploring}
Martin Weyssow, Xin Zhou, Kisub Kim, David Lo, and Houari Sahraoui.
\newblock Exploring parameter-efficient fine-tuning techniques for code generation with large language models.
\newblock {\em arXiv preprint arXiv:2308.10462}, 2023.

\bibitem{wang2022no}
Chaozheng Wang, Yuanhang Yang, Cuiyun Gao, Yun Peng, Hongyu Zhang, and Michael~R Lyu.
\newblock No more fine-tuning? an experimental evaluation of prompt tuning in code intelligence.
\newblock In {\em Proceedings of the 30th ACM joint European software engineering conference and symposium on the foundations of software engineering}, pages 382--394, 2022.

\bibitem{wang2021codet5}
Yue Wang, Weishi Wang, Shafiq Joty, and Steven~CH Hoi.
\newblock Codet5: Identifier-aware unified pre-trained encoder-decoder models for code understanding and generation.
\newblock {\em arXiv preprint arXiv:2109.00859}, 2021.

\bibitem{liu2021self}
Xiao Liu, Fanjin Zhang, Zhenyu Hou, Li~Mian, Zhaoyu Wang, Jing Zhang, and Jie Tang.
\newblock Self-supervised learning: Generative or contrastive.
\newblock {\em IEEE transactions on knowledge and data engineering}, 35(1):857--876, 2021.

\bibitem{radford2021learning}
Alec Radford, Jong~Wook Kim, Chris Hallacy, Aditya Ramesh, Gabriel Goh, Sandhini Agarwal, Girish Sastry, Amanda Askell, Pamela Mishkin, Jack Clark, et~al.
\newblock Learning transferable visual models from natural language supervision.
\newblock In {\em International conference on machine learning}, pages 8748--8763. PMLR, 2021.

\bibitem{tian2020contrastive}
Yonglong Tian, Dilip Krishnan, and Phillip Isola.
\newblock Contrastive multiview coding.
\newblock In {\em Computer Vision--ECCV 2020: 16th European Conference, Glasgow, UK, August 23--28, 2020, Proceedings, Part XI 16}, pages 776--794. Springer, 2020.

\bibitem{you2020graph}
Yuning You, Tianlong Chen, Yongduo Sui, Ting Chen, Zhangyang Wang, and Yang Shen.
\newblock Graph contrastive learning with augmentations.
\newblock {\em Advances in neural information processing systems}, 33:5812--5823, 2020.

\bibitem{brinzea2022contrastive}
Razvan Brinzea, Bulat Khaertdinov, and Stylianos Asteriadis.
\newblock Contrastive learning with cross-modal knowledge mining for multimodal human activity recognition.
\newblock In {\em 2022 International Joint Conference on Neural Networks (IJCNN)}, pages 01--08. IEEE, 2022.

\bibitem{chen2020simple}
Ting Chen, Simon Kornblith, Mohammad Norouzi, and Geoffrey Hinton.
\newblock A simple framework for contrastive learning of visual representations.
\newblock In {\em International conference on machine learning}, pages 1597--1607. PMLR, 2020.

\bibitem{abdollah2023self}
Mohammad~Mahdi Abdollah~Pour, Parsa Farinneya, Armin Toroghi, Anton Korikov, Ali Pesaranghader, Touqir Sajed, Manasa Bharadwaj, Borislav Mavrin, and Scott Sanner.
\newblock Self-supervised contrastive bert fine-tuning for fusion-based reviewed-item retrieval.
\newblock In {\em European Conference on Information Retrieval}, pages 3--17. Springer, 2023.

\bibitem{luo2022clip4clip}
Huaishao Luo, Lei Ji, Ming Zhong, Yang Chen, Wen Lei, Nan Duan, and Tianrui Li.
\newblock Clip4clip: An empirical study of clip for end to end video clip retrieval and captioning.
\newblock {\em Neurocomputing}, 508:293--304, 2022.

\bibitem{zhu2022xlcost}
Ming Zhu, Aneesh Jain, Karthik Suresh, Roshan Ravindran, Sindhu Tipirneni, and Chandan~K Reddy.
\newblock Xlcost: A benchmark dataset for cross-lingual code intelligence.
\newblock {\em arXiv preprint arXiv:2206.08474}, 2022.

\bibitem{yao2018staqc}
Hu~Yao and et~al.
\newblock Staqc: A systematically mined question-code dataset from stack overflow.
\newblock In {\em Proceedings of the World Wide Web (WWW'18) Conference}, pages 135--144, 2018.

\bibitem{bahrami2021pytorrent}
Mehdi Bahrami, NC~Shrikanth, Shade Ruangwan, Lei Liu, Yuji Mizobuchi, Masahiro Fukuyori, Wei-Peng Chen, Kazuki Munakata, and Tim Menzies.
\newblock Pytorrent: A python library corpus for large-scale language models.
\newblock {\em arXiv preprint arXiv:2110.01710}, 2021.

\bibitem{rao2021search4code}
Nikitha Rao, Chetan Bansal, and Joe Guan.
\newblock Search4code: Code search intent classification using weak supervision.
\newblock In {\em 2021 IEEE/ACM 18th International Conference on Mining Software Repositories (MSR)}, pages 575--579. IEEE, 2021.

\bibitem{shi2021semantic}
Chaochen Shi, Yong Xiang, Jiangshan Yu, and Longxiang Gao.
\newblock Semantic code search for smart contracts.
\newblock {\em arXiv preprint arXiv:2111.14139}, 2021.

\bibitem{kairajarvi2020towards}
Sami Kairajärvi, Andrei Costin, and Timo Hämäläinen.
\newblock Isadetect: Usable automated detection of cpu architecture and endianness for executable binary files and object code.
\newblock In {\em Proceedings of the Tenth ACM Conference on Data and Application Security and Privacy}, CODASPY ’20. ACM, March 2020.

\bibitem{guo2024deepseekcoder}
Daya Guo, Qihao Zhu, Dejian Yang, Zhenda Xie, Kai Dong, Wentao Zhang, Guanting Chen, Xiao Bi, Y~Wu, YK~Li, et~al.
\newblock Deepseek-coder: When the large language model meets programming--the rise of code intelligence.
\newblock {\em arXiv preprint arXiv:2401.14196}, 2024.

\bibitem{ganesan2018rouge}
Kavita Ganesan.
\newblock Rouge 2.0: Updated and improved measures for evaluation of summarization tasks.
\newblock {\em arXiv preprint arXiv:1803.01937}, 2018.

\end{thebibliography}

\end{document}